\documentclass[10pt,journal,compsoc]{IEEEtran}
\ifCLASSOPTIONcompsoc
  \usepackage[nocompress]{cite}
\else
  \usepackage{cite}
\fi
\usepackage{amsfonts}
\usepackage{bm}
\usepackage{amsmath}
\usepackage{amssymb}  % assumes amsmath package installed

\DeclareMathOperator*{\argmin}{\arg\,\min}
\DeclareMathOperator*{\Exp}{\mathbb{E}}
\usepackage{multirow}   %使用multirow必须加载该package

\usepackage{microtype}
\usepackage{amsthm}
\newtheorem{theorem}{Theorem}%[section]
\usepackage{subfig}
\usepackage{graphicx}
\usepackage{algorithm}
\usepackage{algorithmic}
\usepackage{url}
\ifCLASSINFOpdf
\else
\fi
\begin{document}
\newcommand{\norm}[1]{\left\lVert#1\right\rVert}
\newcommand{\innerproduct}[2]{\left\langle#1, #2\right\rangle}

\renewcommand{\a}{\mathbf{a}}
\newcommand{\x}{\mathbf{x}}
\newcommand{\y}{\mathbf{y}}
\newcommand{\s}{\mathbf{s}}
\newcommand{\z}{\mathbf{z}}
\newcommand{\f}{\mathbf{f}}
\newcommand{\bmu}{\bm{\mu}}
\newcommand{\bsigma}{\bm{\sigma}}
\newcommand{\bTheta}{\bm{\Theta}}
\newcommand{\bSigma}{\bm{\Sigma}}
\newcommand{\w}{\bm{w}}
\newcommand{\rr}{\bm{r}}
\newcommand{\dd}{\mathbf{d}}
\newcommand{\balpha}{\bm{\alpha}}
\newcommand{\g}{\mathbf{g}}
\newcommand{\e}{\mathbf{e}}
\newcommand{\bx}{\mathbf{X}}
\newcommand{\by}{\mathbf{Y}}
\newcommand{\bz}{\mathbf{Z}}
\newcommand{\be}{\mathbf{E}}
\newcommand{\bl}{\mathbf{L}}
\newcommand{\bs}{\mathbf{S}}
\newcommand{\bg}{\mathbf{G}}
\newcommand{\ba}{\mathbf{A}}
\newcommand{\bc}{\mathbf{C}}
\newcommand{\bb}{\mathbf{B}}
\newcommand{\bu}{\mathbf{U}}
\newcommand{\bq}{\mathbf{Q}}
\newcommand{\bv}{\mathbf{V}}
\newcommand{\bp}{\mathbf{P}}
\newcommand{\bw}{\mathbf{W}}
\newcommand{\bd}{\mathbf{D}}
\newcommand{\bi}{\mathbf{I}}
\newcommand{\bj}{\mathbf{J}}
\renewcommand{\u}{\mathbf{u}}
\renewcommand{\v}{\mathbf{v}}
\newcommand{\p}{\mathbf{p}}
\renewcommand{\b}{\mathbf{b}}
\newcommand{\bflambda}{\mathbf{\lambda}}
\newcommand{\A}{\mathcal{A}}
\newcommand{\X}{\mathbf{X}}
\newcommand{\W}{\mathbf{W}}
\newcommand{\sgn}{\mbox{sgn}}
\newcommand{\diag}{\mbox{diag}}
\newcommand{\armin}{\mbox{argmin}}
\newcommand{\rank}{\mbox{rank}}
\newcommand{\<}{\left\langle}
\renewcommand{\>}{\right\rangle}
\newcommand{\lbar}{\left\|}
\newcommand{\rbar}{\right\|}
\newcommand{\fan}[1]{\Vert #1 \Vert}
\newcommand{\qileft}{[\kern-0.15em[}
\newcommand{\qiLeft}{\left[\kern-0.4em\left[}
\newcommand{\qiright}{]\kern-0.15em]}
\newcommand{\qiRight}{\right]\kern-0.4em\right]}
\renewcommand{\algorithmicrequire}{\textbf{Input:}} % Use Input in the format of Algorithm
\renewcommand{\algorithmicensure}{\textbf{Output:}} % Use Output in the format of Algorithm
\newcommand{\eg}{\emph{e.g. }}
\newcommand{\ie}{\emph{i.e.}}
\newcommand{\wrt}{\emph{w.r.t. }}

\title{Streaming Label Learning for Modeling Labels on the Fly}

\author{Shan~You,~Chang~Xu,~Yunhe~Wang,~Chao~Xu and~Dacheng~Tao,~\IEEEmembership{Fellow,~IEEE}% <-this % stops a space
\IEEEcompsocitemizethanks{\IEEEcompsocthanksitem M. Shell was with the Department
of Electrical and Computer Engineering, Georgia Institute of Technology, Atlanta,
GA, 30332.\protect\\
E-mail: see http://www.michaelshell.org/contact.html
\IEEEcompsocthanksitem J. Doe and J. Doe are with Anonymous University.}% <-this % stops an unwanted space
\thanks{Manuscript received April 19, 2005; revised August 26, 2015.}}


\markboth{Journal of \LaTeX\ Class Files,~Vol.~14, No.~8, August~2015}%
{Shell \MakeLowercase{\textit{et al.}}: Bare Demo of IEEEtran.cls for Computer Society Journals}



\IEEEtitleabstractindextext{%
\begin{abstract}
It is challenging to handle a large volume of labels in multi-label learning. However, existing approaches explicitly or implicitly assume that all the labels in the learning process are given, which could be easily violated in changing environments. In this paper, we define and study streaming label learning (SLL), \ie labels are arrived on the fly, to model newly arrived labels with the help of the knowledge learned from past labels.  The core of SLL is to explore and exploit the relationships between new labels and past labels and then inherit the relationship into hypotheses of labels to boost the performance of new classifiers. In specific, we use the label self-representation to model the label relationship, and SLL will be divided into two steps: a regression problem and a empirical risk minimization (ERM) problem. Both problems are simple and can be efficiently solved. We further show that SLL can generate a tighter generalization error bound for new labels than the general ERM framework with trace norm or Frobenius norm regularization. Finally, we implement extensive experiments on various benchmark datasets to validate the new setting. And results show that SLL can effectively handle the constantly emerging new labels and provides excellent classification performance.
\end{abstract}

}

\maketitle

\IEEEdisplaynontitleabstractindextext

% \fi
\IEEEpeerreviewmaketitle

\IEEEraisesectionheading{\section{Introduction}\label{sec:introduction}}

\IEEEPARstart{M}{ulti-label} learning has a great many achievements and prospects for its successful application to real-world problems, such as text categorization \cite{yang2009effective,li2015supervised}, gene function classification \cite{barutcuoglu2006hierarchical,cesa2012synergy} and image/video annotation \cite{wang2008automatic,cabral2011matrix}. In the multi-label learning problem, each example can be associated with multiple and nonexclusive labels, and the goal of learning is to allocate the most relevant subset of labels to a new example. In the era of big data, the size of label set is constantly increasing. For example, there are already millions of image tags in Flickr and categories in Wikipedia. Hence, the research challenge is to design scalable yet effective multi-label learning algorithms, which are capable to compromise the conflict between the prodigious number of labels and the limited computation resource.

A straightforward approach for multi-label learning is 1-vs-all or Binary Relevance (BR) \cite{tsoumakas2010mining}, which learns an independent classifier for each label. However, the constant increase on the size of label set makes it computationally infeasible. The prevalent technique to deal with label proliferation problem is to \textit{shrink} the large label space by embedding original high-dimensional label vectors into low-dimensional representations. Different projection mechanisms can be adopted for transforming label vectors, including compressed sensing \cite{hsu2009multi}, principal component analysis \cite{tai2012multilabel}, canonical correlation analysis \cite{zhang2011multi}, singular value decomposition \cite{chen2012feature} and Bloom filters \cite{cisse2013robust}. The predictions made in the low-dimensional label space are then transformed  back onto the original high-dimensional label space via a decomposition matrix \cite{yu2014large,sun2014multi} or k-nearest neighbor (kNN) technique \cite{bhatia2015sparse}. Additionally, some works \cite{balasubramanian2012landmark,bi2013efficient} attempt to select a small yet sensible subset of labels to represent the entire label set, and then learn hypotheses regrading this smaller label set.

Aforementioned learning methods successfully remedy the label proliferation problem and have achieved promising performance in different multi-label tasks. However, these methods may be restricted in two aspects. (a) Nearly all of these algorithms explicitly or implicitly assume that \emph{all} the labels in the learning process are given for once, and thus they can only tackle a \textit{static label setting}, which could be easily violated in changing environments. In practice, there is a rapid increase in the volume of labels as the understanding of data goes more in depth, disabling the static label setting in consequence. For example, in social network, users usually belong to different groups or clubs according to their individual characteristics or interests. With the fast development of information techniques and the convenient information transmission, there could emerge new interest groups or clubs, which should be timely and accurately recommended to prospective members. In event detection problem, it is urgent to timely and effectively investigate an emerging new event, which is excluded in the early detection systems. In this way, we can immediately integrate new events into the previous detection system by borrowing the knowledge from past events. Therefore, involvement of constantly emerging labels is very significant for the multi-label learning. (b) Although there are some tricks to adapt classical multi-label algorithms to handle emerging new labels, they could have various disadvantages. More precisely, independently learning for new labels would neglect the knowledge harvested from past labels; integrating new labels and past labels to re-train a new multi-label model requires a huge computation cost, which thus decreases the scalability of the multi-label system, especially when dealing with large scale scenario. As a result, it is challenging to efficiently and accurately model the emergence of the new labels.

Targeting at the both problematic aspects, we define and study streaming label learning (SLL), \ie, labels are arrived on the fly, to learn a model for the newly-arrived $k~(\geq 1)$ labels given a well-trained model for the existing $m$ (usually large) labels within multi-label learning context. The proposed streaming label learning algorithm equips with the capability of modelling newly-arrived labels with the help of the knowledge learned from past labels to timely and effectively respond to the changing and demands of environments. The core of SLL is to explore and exploit the relationships between new labels and past labels. Instead of decomposing the label matrix into label vectors in terms of different data points \cite{tsoumakas2010mining,yu2014large,bhatia2015sparse}, we examine it from the perspective of label space and represent each label through the response values on examples. Based on the idea of ``labels represent themselves", the label structure exploited for labels self-representation stands for relationships between labels, which can be inherited by hypotheses of labels as well. Given the relationships between past labels and newly arrived labels, we can thus easily model the new labels with the help of the well-trained multi-label model on the large number of past labels. We theoretically suggest that the generalization ability of hypotheses of newly arrived labels can be largely improved with the knowledge harvested from past labels. Experimental results on large-scale real-world datasets demonstrate the significance of studying streaming label learning and the effectiveness of the proposed algorithm in timely and effectively learning new labels.

The rest of the paper is organized as follows. In Section 2, we formulate the streaming label learning problem and propose the corresponding mathematical model. The optimization process is elaborated in Section 3 and theoretical analysis is given in Section 4. In Section 5, we present and analyze the experimental results, with concluding remarks stated in Section 6. All detailed proofs are shown in Appendix section.

%\section{Related Work}

\section{Problem Formulation}
In this section we present a streaming label learning mechanism to handle the emerging new labels with the help of the knowledge learned from past labels, which is able to explore the previously well-trained multi-label model over a large number of labels and get rid of intensive computation cost. The proposed algorithm seeks to exploit label relationship via label self-representation, which has an important influence on the hypotheses of labels.

We first state multi-label learning (MLL) and introduce frequent notations. Let the given training data set denoted by $\mathcal{D}=  \{(\x_1,\y_1),...,(\x_n,\y_n)\}$, where $\x_i\in\mathcal{X}\subseteq\mathbb{R}^d$ is the input feature vector and $\y_i\in\mathcal{Y}\subseteq\{-1,1\}^m$ is the corresponding label vector. Moreover, $y_{ij}=1$ iff the $j$-th label is assigned to the example $\x_i$ and $y_{ij} = -1$ otherwise. Let $\bx=[\x_1,...,\x_n]\in\mathbb{R}^{d\times n}$ be the data matrix and $\by=[\y_1,...,\y_n]\in\{-1,1\}^{m\times n}$ be the label matrix. Given dataset $\mathcal{D}$, multi-label learning aims to learn a function $f: \mathbb{R}^d \rightarrow \{-1,1\}^m$ that generates the prediction on label vector for a test point.

\subsection{Label Relationship}\label{section2.1}
Probe of label relationship is demonstrated to be critical and beneficial in boosting the performance of multi-label learning \cite{bi2011multi,cheng2010bayes,hariharan2010large}. Given the label matrix $\by=[\y_1,...,\y_n]\in\{-1,1\}^{m\times n}$, where $Y_{ij}$ indicates the response of $i$-th label on example $\x_j$, most works  \cite{tsoumakas2010mining,yu2014large,bhatia2015sparse} treat $\by$ from the perspective of column (example) and investigate different techniques to transform these example-wise vectors. By contrast, we propose to examine the label matrix from the label perspective, namely, considering the row-wise vectors as the representations of different labels. This thus enables us to specify the abstract concept of a label via its responses on $n$ examples. To facilitate the mathematical notations, we equivalently examine the columns of $\by$'s transpose, denoted as $\by^* = \by^T = [\y^*_1,...,\y^*_m]\in\{-1,1\}^{n\times m}$.

In the following, we proceed to introduce two important assumptions for streaming label learning problem.

\textbf{$\bullet$ Label Self-representation.} Given $m$ labels indexed by $l_i(i=1,...,m)$, each of them can be represented by vectors $\{\y_1^*,...,\y_m^*\}$. We employ a valuable assumption of ``labels represent themselves" to model the label relationship. Specifically, a label is assumed to be represented as a combination of other labels. For example, a linear representation is utilized for a given label $l_i$,
 \begin{equation}\label{label}
\y_i^* = \sum_{j\neq i}s_{i}^j\y_j^*
\end{equation}
where $\s_i = [s_i^1,...,s_i^m]^T$ is the coefficient vector to reconstruct label $l_i$ and $s_i^i=0$ excludes $l_i$ itself in reconstruction. Moreover, if $s_i^j>0$, then label $l_j$ has positive influence on label $l_i$ in Eq.(\ref{label}), while  $s_i^j<0$ implies that label $l_j$ has negative influence on label $l_i$. $\s_i$ is encouraged to be sparse, so that label $l_i$ only has connections with several labels. 

\textbf{$\bullet$ Hypotheses of Labels.} Multi-label learning aims to learn better hypotheses of labels with the help of relationship between labels. A simple yet effective approach to formulate the process of multi-label decision $\mathbb{R}^d \rightarrow \{-1,1\}^{m\times n}$ is via using function $f(\x;W) = W^T\x = [\w_1^T\x,...,\w_m^T\x]^T$. The multi-label classifier $W$ can thus be regarded as the composition of classifiers regarding different labels $\{\w_1,...,\w_m\}$, where $\w_i$ is the classifier \wrt label $l_i$. We assume that the relationship between labels can be inherited by classifiers of different labels. Given label $l_i$ represented by its related labels $\mathcal{N}_i = \{l_j|s_i^j\neq0,j=1,...,m\}$ according to  Eq.(\ref{label}), classifier $\w_i$ w.r.t. label $l_i$ can thus be represented by the classifiers regarding those related labels using the same coefficient vector $\s_i$,
\begin{equation}\label{weight}
\w_i = \sum_{j\neq i}s_{i}^j\w_j.
\end{equation}
Broadly speaking, label relationship acts as a regularization of multi-label classifier $W$, which encourages $W$ to be represented by itself as well.

The linear self-representation of labels is a simple yet effective assumption within multi-label learning indeed. There usually exist significant dependencies among labels in multi label learning. With the extension of real datasets, these dependencies are enhanced as well. Thus it is easy for some specific label to investigate a group of ``neighborhood" labels involved in its linear representation among a great many labels. \footnote{We will validate the linear self-representation of labels empirically in Section 6.} Note that our operation over label matrix resembles label selection techniques in \cite{balasubramanian2012landmark}, which also assumes linear self-representation of labels, but they are developed from distinct perspectives. Label selection focuses on selecting a shared subset of labels to recover all the given labels. Nevertheless, the proposed label self-representation aims to accurately represent the label current in progress, which will not be distracted by the reconstruction results of other labels. Besides, we propagate the label relationship exploited through label self-representation into the process of learning multi-label classifiers, instead of independently learning classifiers for the selected label subset. It is instructive to note that the label self-representation operation implicitly encourages $W$ to be low rank, which is also a widely-used assumption in multi-label learning \cite{sun2014multi,yu2014large}.

\subsection{Streaming Label Learning}
Conventional well-trained multi-label learning model on data associated with a large number of labels is difficult to be adapted to the newly arrived labels without computationally intensive re-training, and the label relationship discovered by existing methods cannot be straightforwardly extended with the emerging new labels as well.

This section details the proposed steaming label learning (SLL) mechanism, designed for accommodating emerging new labels. Basically, SLL consists of two steps. For the $k$ newly arrived labels, we first exploit their relationships between $m$ past labels, and then learn their corresponding hypotheses with the help of previously well-trained model regarding past labels and the exploited label relationship.

We first assume there is only one newly-arrived label $(k=1)$ and then extend it into a mini-batch setting ($k\geq 1$). Denote a well-trained multi-label learning model over $m$ labels as $W_m=[\w_1,...,\w_m]$, and the label matrix of $n$ examples aligned in label dimension is denoted as $Y_m^*=[\y_1^*,...,\y_m^*]$. Besides, given matrix $S_m$ describing label relationship, we then have $Y^*_m \approx Y^*_m S_m$ and $W_m \approx W_m S_m$.

\textbf{Streaming label learning with one label.} Given a new label $l_{m+1}$ represented by the response vector $\y_{m+1}^*$ on $n$ examples, we assume that it can be represented by the past $m$ labels $\y^*_{m+1} = \sum_{j=1}^m s_{m+1}^j \y^*_j$, where $\s_{m+1}$ is the coefficient vector of the new label $l_{m+1}$, and can be determined by solving the following optimization problem:
\begin{equation}\label{new:weight}
\argmin_{\s_{m+1}\in\mathbb{R}^m} \quad\frac{1}{2}\norm{\y^*_{m+1}-Y^*_m\s_{m+1}}_2^2 + \lambda \norm{\s_{m+1}}_1,
\end{equation}
where the least squares acts as a residual penalty for the label representation and $\lambda>0$ encourages sparsity. In this way, we can obtain the representation $\s_{m+1}$ of the new label $l_{m+1}$.

The label relationship between the new label and past labels can provide us helpful information to learn the hypothesis regarding the new label. According to hypotheses of labels in Eq.(\ref{weight}), we have
\begin{equation}\label{label1}
\w_{m+1} = W_m \s_{m+1}
\end{equation}
for new label $l_{m+1}$. Eq.(\ref{label1}) actually provides prior information for the new classifier $\w_{m+1}$ to be learned. By further considering its prediction error, $\w_{m+1}$ can be learned by minimizing the following objective function:
\begin{multline}\label{new:classifierobj}
J(\w_{m+1}) = \sum_{i=1}^n\ell(y^*_{m+1,i},\x_i^T\w_{m+1})\\ + \frac{\beta}{2}\norm{\w_{m+1}-W_m \s_{m+1}}_2^2
\end{multline}
where $\beta>0$ is a regularization parameter and $\ell(\cdot,\cdot)$ is a loss function to measure the discrepancy between the ground-truth label and the prediction. We choose the $\ell_2$ loss function in our experiment for simplicity though our SLL can adapt to other loss functions, since $\ell_2$ loss is shown to stand out in most cases of multi-label classification tasks comparing to other loss functions, such as logistic loss and $L_2$-hinge loss \cite{yu2014large}. Therefore, the new classifier can be learned subsequently, but integrated with the already-learned knowledge of past labels.

SLL with one new label can be naturally extended into a mini-batch setting, where a mini-batch of new labels instead of one single label is processed at a time.

\textbf{Mini-batch extension.} Given a batch of $k$ new labels $l_{new} = \{l_{m+1},...,l_{m+k}\}$ represented by vectors $Y^*_{new} = \{\y^*_{m+1},...,\y^*_{m+k}\}$, the challenging part is that we need to consider not only the relationships between new labels and past labels, but also those among new labels.

Suppose that each new label is reconstructed with the help of all the other labels (new labels and past labels), i.e. $l_{m+i} = \sum_{j=1}^{m+k} s_{m+i}^j l_j ,j\neq m+i$. According to Eq.(\ref{label}), we can obtain
\begin{equation}
Y^*_{new} = Y^*_{m+k}S_{new}
\end{equation}
with $Y^*_{m+k} = [Y^*_m,Y^*_{new}]$. Then the representation of new labels can be also solved through the following optimization problem as Eq.(\ref{new:weight}),
\begin{equation}\label{online:weight}
\begin{array}{rl}
\arg\displaystyle{\min_{S_{new}}} & \frac{1}{2}\norm{Y^*_{new} - Y^*_{m+k}S_{new}}_F^2+ \lambda\norm{S_{new}}_{1,1}\\
s.t.& (S_{new})_{m+i,i} = 0,\ \forall i = 1,...,k.\\
\end{array}
\end{equation}
where $(S_{new})_{m+i,i}=0$ is to exclude each individual label from its reconstruction. As a result, the representation of new labels can be obtained. Moreover, considering the partitioning of $S_{new} = [S_{new}^{(1)};S_{new}^{(2)}]$, we have $Y^*_{new} = Y^*_{m+k}S_{new} = Y^*_mS_{new}^{(1)} + Y^*_{new}S_{new}^{(2)}$. Thus we can observe that $S_{new}^{(1)}$ corresponds to the representation from $m$ past labels while $S_{new}^{(2)}$ means the interactive representation of the $k$ new labels, which coheres with the assumption we make.

After obtaining $S_{new}$, it can also be employed in the process of learning new classifiers, $W_{new} = [W_m,W_{new}]S_{new} = W_mS_{new}^{(1)} + W_{new}S_{new}^{(2)}$ where $W_{new}$ is the parameter matrix for new labels. Then the optimization function is formulated as
\begin{multline}\label{new:batch:classifier}
\tilde{J}(W_{new}) = \frac{1}{2}\sum_{i=1}^n\norm{\y_{new}^{i}-W_{new}^T\x_i}_2^2\\
+  \frac{\beta}{2}\norm{W_{new}(\bi-S_{new}^{(2)})-W_mS_{new}^{(1)}}_F^2
\end{multline}
Note that the adopted loss function in Eq.(\ref{new:batch:classifier}) is decomposable, namely, $\norm{\y_{new}^{i}-W_{new}^T\x_i)}_2^2 = \sum_{j=1}^k (y_{new}^{ij}-\w_{m+j}^T\x_i)^2$ where $y_{new}^{ij}$ is the $j$-th new label value of the $i$-th example $\x_i$. Besides, when $S_{new}^{(2)}=\bf{0}$, it means that we neglect the relationships within new labels, and only investigate the relationships between new labels and past labels. This case coheres with the single new label scenario, and thus it can be viewed as a special case with batch size 1.

The framework of the proposed algorithm is summarized in Algorithm \ref{alg:online}.

\begin{algorithm}[tb]
   \caption{Framework of Streaming Label Learning}
   \label{alg:online}
\begin{algorithmic}[1]
\REQUIRE{A well-trained multi-label classifier $W\in\mathbb{R}^{d\times m}$ with respect to $m$ (usually large enough) labels.}
\WHILE{areMoreLabelsAvailable() }
\STATE $\{(\x_i,\y_i)\}_{i=1}^n\leftarrow$ getNewLabelTrainingData()
\STATE $Y^*_{m+k} = [Y_m^*,Y_{new}^*] \leftarrow \by^T=[\y_1,...,\y_n]^T$
\STATE Label representation structure $S_{new}$: solving Eq.\ref{online:weight},\\
\STATE New labels classifier $W_{new}$: solving Eq.\ref{new:batch:classifier}.\\
\STATE $W \leftarrow [W,W_{new}]$.
\STATE $m \leftarrow m +k$
\ENDWHILE
\ENSURE{multi-label classifier with parameter $W$ in terms of all $m$ labels.}
\end{algorithmic}
\end{algorithm}

\section{Optimization}
In this section, we show the details of optimization in SLL. Basically, the optimization mainly consists of two parts, \ie solving the label representation $S$ and the classifier parameter matrix $W$. Additionally, we introduce a natural method to initialize the proposed steaming label learning algorithm.

\subsection{Optimizing the New Label Representation $S$}
For a single new label (see Problem (\ref{new:weight})), the optimization is unconstrained yet with a non-smooth regularization term. In fact, it is an $\ell_1$-regularized linear least-squares problem or Lasso. This problem has been thoroughly investigated in many literatures, and there exist a great many algorithms to solve it efficiently, such as alternating direction method of multipliers (ADMM) \cite{boyd2011distributed}, least angle regression (LARS) \cite{efron2004least}, grafting \cite{perkins2003online} and feature-sign search algorithm \cite{lee2006efficient}. Also there are some off-the-shelf toolboxes or packages solving it, \eg~CVX \cite{cvx,gb08}, TFOCS \cite{becker2011templates} and SPAMS \cite{mairal2009online}. Note that the optimization of Problem (\ref{new:weight}) depends on the dimension of the label set. To handle the label proliferation problem, we propose to implement clustering trick \cite{bhatia2015sparse} to all labels, then select labels with an appropriate number for each cluster and compose a relatively small label dictionary in order to improve the efficiency.

As for new labels in the mini-batch setting, Problem (\ref{online:weight}) can be efficiently solved over each column of $S_{new}$ using parallel techniques. And each subproblem can also be handled like Eq.(\ref{new:weight}).

\subsection{Optimizing the Classifier $W$ for New Labels }

\textbf{$\w_{m+1}$ in Problem (\ref{new:classifierobj})}. Various gradient-based or subgradient-based methods can be adopted to minimize Eq.(\ref{new:classifierobj}). Different from \cite{yu2014large}, which seeks to process a large number of labels all at once and has to turn to cheap methods, such as conjugate gradient (CG) method, Eq.(\ref{new:classifierobj}) provides a practical solution to learn multiple labels in a streaming manner, and thus largely alleviates the challenge of a prodigious number of labels on machine load and computational cost. We only concern a $d$-dimension optimization problem, which enables us to adopt more accurate methods, like LBFGS search, with little computational cost. With the squared loss, we can even obtain a closed-form solution of Eq.(\ref{new:classifierobj}),
\begin{equation}\label{eq:solution:l2}
\w_{m+1} \leftarrow (\bx\bx^T+\beta\bi)^{-1} (\bx\y_{m+1}^*+\beta W_m \s_{m+1}).
\end{equation}
which only needs to inverse a $d\times d$ matrix instead of $dr\times dr$ in \cite{yu2014large}, where $r$ is the low rank upper bound.

\textbf{$W_{new}$ in Problem (\ref{new:batch:classifier})}. Similarly, Eq.(\ref{new:batch:classifier}) can also be solved by various gradient-gradient methods. However, the bottleneck is the calculation of gradients and the corresponding Hessian matrix for its cost may be extremely large. Let $\w_{new} = vec(W_{new})\in\mathbb{R}^{dk}$, where $vec(\cdot)$ is the vectorization of a matrix. Since loss function $\ell$ is separable over each $\w_{m+i}$, then the gradient and Hessian matrix of $\w_{new}$ can be calculated using the gradient and Hessian matrix over each $\w_{m+i}$.
Besides, denoting $D_{new} = ((\bi-S_{new}^{(2)})\otimes\bi)$ and $\z_{new} = vec(W_mS_{new}^{(1)})$, the residual penalty $\frac{1}{2}\norm{W_{new}(\bi-S_{new}^{(2)})-W_mS_{new}^{(1)}}_F^2$ can be rewritten as $\frac{1}{2}\norm{D_{new}^T\w_{new}-\z_{new}}_2^2$, whose gradient and Hessian matrix are easy to calculate. Let $\ell'(a,b) = \frac{\partial}{\partial b}\ell(a,b), \ell''(a,b) = \frac{\partial^2}{\partial b^2}\ell(a,b)$, then the gradient and Hessian-vector multiplication of $\w_{new}$ are:
%\begin{align}
%&\nabla \tilde{J}(\w_{new}) = stack_{j=1}^k[\sum_{i=1}^n\ell'(y_{new}^{ij},\w_{m+j}^T\x_i)\x_i] \nonumber\\
%&\qquad\qquad\qquad+ \beta D_{new}(D_{new}^T\w_{new}-\z_{new})\nonumber\\
%=&vec(\bx G + \beta[W_{new}(\bi-S_{new}^{(2)})-W_mS_{new}^{(1)}](\bi-S_{new}^{(2)})^T)\label{jiandan1}\\
%&\nabla^2 \tilde{J}(\w_{new})\z = \sum_{j=1}^k\sum_{i=1}^n \ell''(y_{new}^{ij},\w_{m+j}^T\x_i)\x_i\x_i^T\z_j\nonumber\\
% &\qquad\qquad\qquad\quad+ \beta D_{new}D_{new}^T\z \nonumber\\
% & = vec(\bx H+\beta Z(\bi-S_{new}^{(2)})(\bi-S_{new}^{(2)})^T)\label{jiandan2}
%\end{align}
\begin{align}
&\nabla \tilde{J}(\w_{new}) = \mbox{stack}\left\{\sum_{i=1}^n\ell'(y_{new}^{ij},\w_{m+j}^T\x_i)\x_i\right\}_{j=1}^k \nonumber\\
&\qquad\qquad\qquad+ \beta D_{new}(D_{new}^T\w_{new}-\z_{new})\nonumber\\
&=vec(\bx G + \beta[W_{new}(\bi-S_{new}^{(2)})-W_mS_{new}^{(1)}](\bi-S_{new}^{(2)})^T)\label{jiandan1}\\
&\nabla^2 \tilde{J}(\w_{new})\z = \mbox{stack}\left\{\sum_{i=1}^n \ell''(y_{new}^{ij},\w_{m+j}^T\x_i)\x_i\x_i^T\z_j\right\}_{j=1}^k\nonumber\\
 &\qquad\qquad\qquad\quad+ \beta D_{new}D_{new}^T\z \nonumber\\
 & = vec(\bx H+\beta Z(\bi-S_{new}^{(2)})(\bi-S_{new}^{(2)})^T)\label{jiandan2}
\end{align}
where $G_{ij} = \ell'(y_{new}^{ij},\w_{m+j}^T\x_i)$, $\z = vec(Z) = vec([\z_1,...,\z_k])$, $H_{ij} = \ell''(y_{new}^{ij},\w_{m+j}^T\x_i)\x_i^T\z_j$ and stack$\{\cdot\}$ means stacking the vectors in the set to form a longer vector according to the ascending index order. As a result, the calculation of gradient and Hessian-vector multiplication can be efficiently obtained using Eqs.(\ref{jiandan1})-(\ref{jiandan2}). For the $\ell_2$ loss function, the key factors $G$ and $H$ in Eq.(\ref{jiandan1})-(\ref{jiandan2}) can be more easily calculated,
\begin{equation}\label{GH}
G = \bx^T W_{new} - Y_{new}^*;\quad H = \bi.
\end{equation}
In SLL, $k$ is usually small and thus we can turn to more refined techniques, such as various line search. However, since the size of Problem (\ref{new:batch:classifier}) is $d\times k$, we propose to adopt cheap methods, such as Conjugate Gradient (CG), when feature dimension $d$ is very large.

\subsection{An Initialization proposal of $W_m$ for Past Labels}
Optimization and implementation of SLL requires an already well-trained model $W_m$ for past $m$ labels, which is critical for the performance over $k$ new labels. In real application, we just utilize an obtained $W_m$ as the input of SLL to model $k$ new labels, without retraining the $m+k$ labels. 

Furthermore, although SLL is designed for modelling new labels, one additional merit is that it also provides a solution for the memory limited multi label learning problem. Almost all existing MLL methods need to load the whole feature and label data into the memory, and they may be fairly restrictive to be trained on low-end computation devices at hand since the datasets are tending to be larger and larger. Fortunately, due to the separate two steps of SLL, we can load label data and feature data successively into memory, since they dominate the memory overhead of both two steps respectively. In this way, SLL can basically decrease half of memory need for the training of MLL, together with considering the dependencies among labels.

In this case, based on the two assumptions in Section \ref{section2.1}, one practical proposal for learning the initial classifier $W_m$ over past $m$ labels is in a similar approach as that for SLL, by minimizing the following objective:
\begin{multline}\label{obj:classifierobj}
\mathcal{J}(W_m,S_m) = \frac{1}{2}\sum_{i=1}^n\norm{\y_i-W_m^T\x_i}_2^2 + \lambda_1\norm{S_m}_{1,1} \\+ \frac{\lambda_2}{2}\norm{W_m-W_mS_m}_F^2+\frac{\lambda_3}{2}\norm{Y^*_m-Y^*_mS_m}_F^2
\end{multline}
where $\norm{S_m}_{1,1}=\sum_{i=1}^m\norm{\s_i}_1$ promotes the sparsity of $S_m = [\s_1,...,\s_m]\in\mathbb{R}^{m\times m}$. $\lambda_i>0~(i=1,2,3)$ are the weight parameters. Usually, we expect that the objective function is minimized given a small reconstruction error of labels, thus much weight (\ie~large $\lambda_3$) should be imposed on $\frac{1}{2}\norm{Y^*_m-Y^*_mS_m}_F^2$ term in Eq.(\ref{obj:classifierobj}).

Problem (\ref{obj:classifierobj}) is basically solved with the alternative iteration strategy, i.e. fixing one variable and optimizing the other until convergence. As for solving for $S_m$, it is still a lasso problem, and can adopt the same methods in solving Problem (\ref{online:weight}). As for solving $W_m$, it can be viewed as a special case of Problem (\ref{new:batch:classifier}) with $S_{new}^{(1)} = \bm{0}$, thus Eq.(\ref{new:batch:classifier}) is of size $d\times m$. In this case, especially referred to large scale labels, we may perform cheap updates and obtain a good approximate solution. For example, Conjugate Gradient (CG) can be employed to significantly reduce the computational complexity based on Eq.(\ref{jiandan1}) and (\ref{jiandan2}).

%\begin{algorithm}[tb]
%   \caption{Optimization of Problem \ref{online:classifier}}
%   \label{alg:classifier}
%\begin{algorithmic}[1]
%\REQUIRE{training data $\bx=[\x_1,...,\x_n]\in\mathbb{R}^{d\times n}$, new label values $Z_{new}\in\{-1,1\}^{n\times k}$, representation weight matrix $S_{new} = [S_{new}^{(1)};S_{new}^{(2)}]$, classifier parameter $W_m$ corresponding to last $m$ labels .}
%\STATE \textbf{Initialization}: $D \leftarrow(\bi_k-S_{new}^{(2)})^T, E \leftarrow (S_{new}^{(1)})^TW_m^T$
%\STATE \textbf{Warm Restart}: \\$W_{new} \leftarrow IndividualLabelLearner(\bx,Z_{new}^T)$
%\STATE $W\leftarrow W_{new}^T$
%\REPEAT
%\FOR{$i = 1$ to $d$}
%\STATE \small $\tilde{\w}_i\leftarrow\left[
%  \begin{array}{cc}
%    \norm{\x_i^r}_2^2\bi_k & D^T \\
%    D & \bm{0}_{k\times k} \\
%  \end{array}
%\right]^{\dagger}\left[
%                       \begin{array}{c}
%                        \b_i   \\
%                         \e_i \\
%                       \end{array}
%                     \right]$,\\
%where $\b_i = \norm{\x_i^r}_2^2\w_i - (W\bx-Z_{new}^T)(\x_i^r)^T.$
%\STATE $\w_i\leftarrow \tilde{\w}_i(1:k).$
%\ENDFOR
%\UNTIL{convergence}
%\STATE $W_{new}\leftarrow W^T$
%\ENSURE{classifier parameter $W_{new}$ for $k$ new labels.}
%\end{algorithmic}
%\end{algorithm}
\section{Theoretical Analysis}
In this section, we theoretically analyze the proposed SLL, regarding the following two aspects: (a) the generalization of the designed classifier for new labels; and (b) the difference between the classifier parameter matrix obtained with streaming labels and that without streaming labels.

\subsection{Generalization Error Bounds}
We first analyze excess risk bounds for SLL. In particular, we present a generalization error bound for the new classier learned in the steaming fashion. Moreover, we show that under some circumstances, our SLL can give a tighter bound than the common trace norm or Frobenius norm regularization in the ERM framework.

Since SLL focuses on boosting the performance of new labels by exploring the knowledge from past labels, it needs a well-trained multi-label classifier as the initialization input, denoted as $W_{old}\in\mathbb{R}^{d\times m}$. Suppose $k$ new labels are involved at a time, then SLL is implemented upon a data distribution $\mathcal{D} = \mathcal{X}\times\{-1,1\}^k$, where $\mathcal{X}\in\mathbb{R}^d$ is the feature space. Training data contains $n$ points $(\x_1,\y_1),...,(\x_n,\y_n)$, which are sampled i.i.d. from the distribution $\mathcal{D}$, where $\x_i\in\mathcal{X}$ is the feature vector and $\y_i\in\{-1,1\}^k$ is the ground-truth label vector. Our SLL is based on the proposed label self-representation and hypotheses, which can be viewed as a regularization. And the regularized set can be written as $\mathcal{W}:= \{W\in\mathbb{R}^{d\times k},\norm{W-W_{old}S}_F \leq \varepsilon, \norm{S}_{1,1} \leq \lambda\}$, where $S$ is the representation weight matrix of new labels with a sparsity controlling parameter $\lambda$. For simplicity, we just analyze the scenario where $k$ new labels have no interaction in their representation.

Given the obtained training data, SLL learns a classifier $\hat{W}$ by minimizing the empirical risk over the regularized set $\mathcal{W}$, $\hat{\mathcal{L}}(W) = \frac{1}{n}\sum_{i=1}^n \sum_{l=1}^k\ell(\y^l_i,\x_i,\w_l)$ and $\hat{W} \in \arg\min_{W\in\mathcal{W}}\hat{\mathcal{L}}(W)$. Define the population risk of an arbitrary $W$ as $\mathcal{L}(W) = \Exp_{(\x,\y)} \qileft \sum_{l=1}^k\ell(\y^l,\x,\w_l) \qiright$, then the goal is to show the learned $\hat{W}$ possesses good generalization, \ie, $\mathcal{L}(\hat{W})\leq\inf_{W\in\mathcal{W}}\mathcal{L}(W) + \epsilon$. We have the following theorem.
\begin{theorem}\label{theorem:1}
Assume we learn a new predictor $W\in\mathbb{R}^{d\times k}$ in terms of $k$ new labels using streaming label learning formulation  $\displaystyle{\hat{W} = \arg\min_{W\in\mathcal{W}}\hat{\mathcal{L}}(W)}$ over a set of $n$ training points, where $\mathcal{W}:= \{\norm{W-W_{old}S}_F \leq \varepsilon, \norm{S}_{1,1} \leq \lambda\}$. Then with probability at least $1-\delta$, we have
\begin{equation*}
\mathcal{L}(\hat{W})  \leq \inf_{W\in\mathcal{W}}\mathcal{L}(W) + \mathcal{O}\left((\varepsilon+\lambda c)\sqrt{\frac{k}{n}}\right) + \mathcal{O}\left(k\sqrt{\frac{\log\frac{1}{\delta}}{n}}\right)
\end{equation*}
where $c = \norm{W_{old}}_F$ and we presume (w.l.o.g) that $\Exp_{\x} \qileft\norm{\x}_2^2\qiright\leq 1$.
\end{theorem}
According to Theorem \ref{theorem:1}, the key term of the upper bound is the second term and it depends on the value of $(\varepsilon+\lambda c)$. $\varepsilon$ is related to the accuracy of the self-representation of label hypotheses, then if the representation performs accurately, $\varepsilon$ can be sufficiently small. $\lambda$ controls the coefficient sparsity and is usually small. Moreover, considering $\norm{W}_F\leq\varepsilon+\lambda\norm{W_{old}}_F$ in $\mathcal{W}$, thus SLL might provide a small upper bound of $\norm{W}_F$ or sometimes of $\norm{W}_*$, which means the generalization error bound generated by SLL could be tighter than the common Frobenius or trace norm regularization, when $\varepsilon$ and $\lambda$ are sufficiently small. Proof of Theorem \ref{theorem:1} is referred to Appendix \ref{proof:1}.

\subsection{Streaming Approximation Error Bound}
We now investigate whether the classifier matrix learned by SLL is seriously deviated from the one learned under the conventional multi-label learning setting. Precisely, suppose we have $k$ labels and the classifier matrix is $\hat{W}_k$, and then we learned a new label classifier matrix $\hat{\w}$ using SLL. However, without SLL we would learn a classifier corresponding to all $k+1$ labels, denoted as $\hat{W}_{k+1}$. The goal is to estimate the difference between $\hat{W}_{k+1}$ and $[\hat{W}_k,\w]$, which reflects the cost of classifier learned by SLL.

Given the training data $\bx=[\x_1,...,\x_n]\in\mathbb{R}^{d\times n}$ and their label matrix $\by=[\y_1,...,\y_n]\in\{-1,1\}^{(k+1)\times n}$, the classifier parameter matrix $\hat{W}_{k+1}$ is determined in the following optimization:
%\vskip -5mm
\begin{equation}\label{deter}
\hat{W}_{k+1} = \argmin_{W\in\mathbb{R}^{d\times (k+1)}}\sum_{i=1}^n \ell(\y_i,\x_i,W) + \frac{\lambda}{2}\norm{W-WS}_F^2,
\end{equation}
where $S$ is the label structure matrix of all $k+1$ labels. Then we have the following theorem:
\begin{theorem}\label{theorem:2}
Given the training data $\{\bx,\by\}$ of $k+1$ labels, the classifier matrix $\hat{W}_{k+1}$ is determined in Eq.(\ref{deter}). Assuming the first $k$ labels are also learned using Eq.(\ref{deter}) denoted as $\hat{W}_k$, while the $(k+1)$th label is learned under the streaming label learning framework, denoted as $\hat{\w}$, then the following inequality holds,
\begin{multline*}
\norm{\hat{W}_{k+1}-[\hat{W}_k,\w]}_F\leq \frac{2}{\lambda\sigma_1^2(\bi-S)+\sigma_1^2(\bx)}\\
\left(\sqrt{2n\Omega C}+\lambda  \tau\sqrt{\norm{\hat{W}}_F^2+\norm{\hat{\w}}_2^2}\right)
\end{multline*}
where $C = \ell_{2}(\by,\bx,[\hat{W}_k,\w])$ is the least squares loss value of the classifier learned by SLL; constant $\tau=\norm{\bi-S}_F^2$ and $\sigma_1(\cdot)$ indicates the smallest singular value. Moreover, we presume (w.l.o.g) that $\norm{\x_i}_2^2\leq \Omega$ and $\bx$ is of full row rank.
\end{theorem}
As indicated in Theorem \ref{theorem:2}, we present an approximation error bound for $\hat{W}_{k+1}-[\hat{W}_k,\hat{\w}]$ using the least squares loss for simplicity and it shows that the bound is directly controlled by the loss of SLL. The better the SLL learns the classifier (\ie the smaller $C$ is), the tighter the bound will be. In this way, if we focus on SLL with much effort, then the learned classifier would not tend to be unsatisfying. Proof of Theorem \ref{theorem:2} is referred to Appendix \ref{proof:2}.

\section{Experimental Results}
In this section, we conduct experiments on SLL to demonstrate its effectiveness and efficiency in terms of dealing with new labels.

\textbf{Datasets.} We select 5 benchmark multi-label datasets to implement the setting of SLL, including three small datasets (Bibtex, MediaMill and Delicious) and two large datasets (EURlex and Wiki10). See Table \ref{data} for the details of these datasets.
\begin{table}
\footnotesize
\caption{Data statistics. $d$ and $L$ are the number of features and labels, respectively; $\bar{d}$ and $\bar{L}$ are the average number of nonzero features and positive labels in an instance, respectively. \# means the size of a dataset.}\label{data}
%\vskip 0.2cm
\begin{tabular}{l|r|r|r|r|r|r}
  %\hline
  Dataset & $d$ & $L$ & \#train & \#test & $\bar{d}$ & $\bar{L}$ \\ \hline
  Bibtex & 1,836 & 159 & 4,880  & 2,515 & 68.74 & 2.40 \\
  MadiaMill & 120 & 101 & 30,993 & 12,914 & 120.00 & 4.38 \\
  Delicious & 500 & 983 & 12,920 & 3,185 & 18.17 & 19.03 \\
  EURlex & 5,000 & 3,993 & 15,539  & 3,809 & 236.69 & 5.31 \\
  Wiki10 & 101,938 & 30,938 & 14,146 & 6,616  & 673.45  & 18.64  \\
%  DeliLarge &782,585 & 205,443 & 196,606  & 100,095 & 301.17 & 75.54 \\
  \hline
\end{tabular}
\end{table}

\textbf{Baseline methods.} We compare the proposed SLL with three competing methods:\\
1). BR \cite{tsoumakas2010mining}. Since our setting focuses on the streaming labels, to our best knowledge, only it can accommodate new labels without retrain the model related to past labels.\\
2). LEML (low rank empirical risk minimization for multi-label learning) \cite{yu2014large}. Since SLL and LEML are both within ERM framework, we analyze the difference of their classification performance.\\
3). SLEEC (sparse local embeddings for extreme classification) \cite{bhatia2015sparse}. Since SLL aims at handling new labels and is capable of scaling to large datasets, we choose the state-of-the-art SLEEC in extreme classification.

\textbf{Evaluation Metrics.} We use three prevalent metrics to measure the performance of all competing methods including our SLL: (a) Hamming loss, which concerns the holistic classification accuracy over all labels, (b) precision at $k$ (P@$k$), which is usually used in tagging or recommendation and only top $k$ predictions are involved in the evaluation, and (c) average AUC, which reflects the ranking performance.

For Lasso-style Problems \ref{new:weight} and \ref{online:weight}, we use a Cholesky-based implementation of the LARS-Lasso algorithm to efficiently solve them with a high accuracy, supported by SPAMS optimization toolbox \cite{mairal2009online}, and implement it in a parallel way. As for obtaining the classifier defined in Problems (\ref{new:classifierobj}) and (\ref{new:batch:classifier}), we propose to use LBFGS line search in small datasets (or use Eq.(\ref{eq:solution:l2}) with moderate $d$ accelerated by GPU) and utilize conjugate gradient descent in large datasets, based on techniques in Eqs.(\ref{jiandan1}) and (\ref{jiandan2}).

%Since SLL handles labels in the streaming fashion, and is capable of dealing with new labels without expensive retraining process, we evaluate its performance from both effectiveness and efficiency aspects.
\begin{figure*}[!ht]
\centering
%\hspace{-10mm}
\subfloat[]
{\includegraphics[width=0.6\columnwidth]{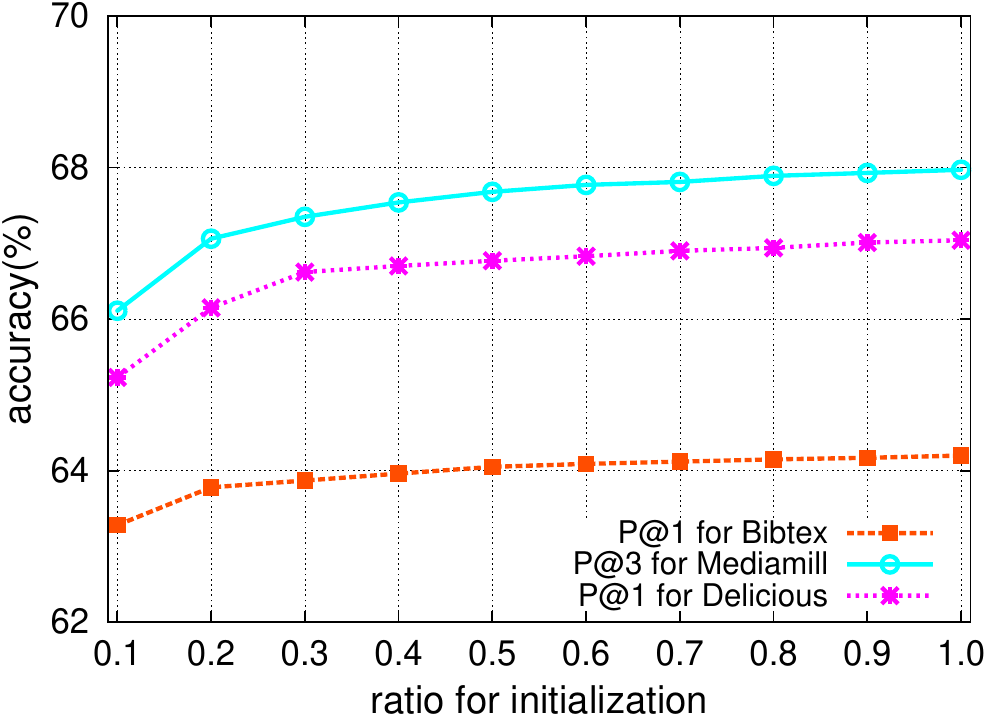}}
\hspace{0.5cm}
\subfloat[]
{\includegraphics[width=0.6\columnwidth]{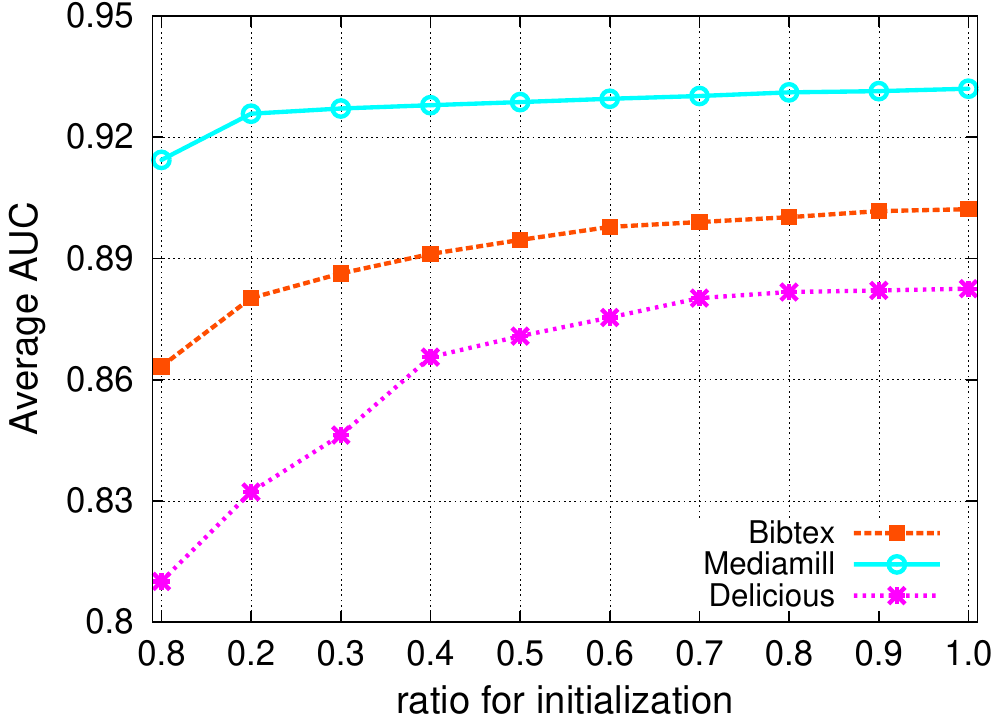}}
\hspace{0.5cm}
\subfloat[]
{\includegraphics[width=0.6\columnwidth]{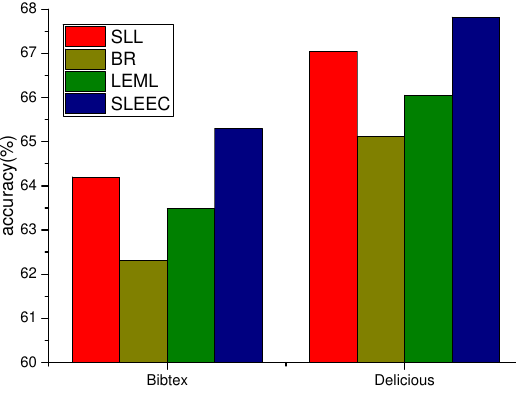}}
%\vskip -4mm
\caption{(a) Accuracy of P@$1$ (Bibtex \& Delicious) and P@$3$ (Mediamill) of SLL in different initial ratio of label size. (b) The average AUC accuracy of SLL in different initial ratio of label size. (c) Comparison of P@$1$ accuracy in all competing methods with 100\% initial ratio.}
\label{fig1}
\end{figure*}
\subsection{Validation of Effectiveness}
%In SLL, streaming labels can be handled in single fashion or in mini-batch fashion. We present the results of both cases respectively.

\textbf{Single new label.}
Basically SLL relies on an initial multi-label classifier related to the past labels, which can be learned by solving Problem (\ref{obj:classifierobj}). We first compare the classification results with varying label ratios which are involved in the initialization training.  Since the streaming labels are processed one by one, only BR can handle this case while other methods (including LEML and SLEEC) have to implement retraining many times, but have identical results with all labels ( ratio = 100\%) involved in initialization. Figure \ref{fig1} shows the P@$3$ accuracy and average AUC results, together with their trends, in various initial label ratios. From these results, we have the following observations. 1) For SLL, more labels involved in the initialization tend to improve its performance. However, with the increase of initial ratio, the improvement would be less obvious; when the ratio is larger than an appropriate value (\eg 70\% for Bibtex, 60\% for Delicious), the performance would keep relatively stable, even sometimes getting worse. Thus selecting the initial label size is very critical when dealing with large datasets, and 50\%$\sim$70\% would be a satisfying option. 2) In terms of all labels (100\% ratio), SLL can yield better accuracies than those of BR and LEML and is competitive with SLEEC. This indicates our used label structure tends to be a stronger regularization in training multi-label classifiers compared to the common Frobenious or trace norm regularization.\\
%\vskip -0.8cm

%
%\begin{figure}[htb]
%\begin{minipage}[t]{0.33\linewidth}
%\centering
%\includegraphics[width=0.8\textwidth]{ys_1}
%\caption{Skyline example of hotel}
%\label{skyline}
%\end{minipage}
%\begin{minipage}[t]{0.33\linewidth}
%\centering
%\includegraphics[width=0.8\textwidth]{ys_2}
%\caption{Skyline example of hotel}
%\label{skyline}
%\end{minipage}
%\begin{minipage}[t]{0.33\linewidth}
%\centering
%\includegraphics[width=0.8\textwidth]{bijiao_accuracy}
%\caption{Skyline example of hotel}
%\label{skyline}
%\end{minipage}
%\end{figure}
%
%

%\vskip -10mm

\begin{figure*}[ht]
\centering
%\hspace{-10mm}
\subfloat[Delicious]
{\includegraphics[width=0.6\columnwidth]{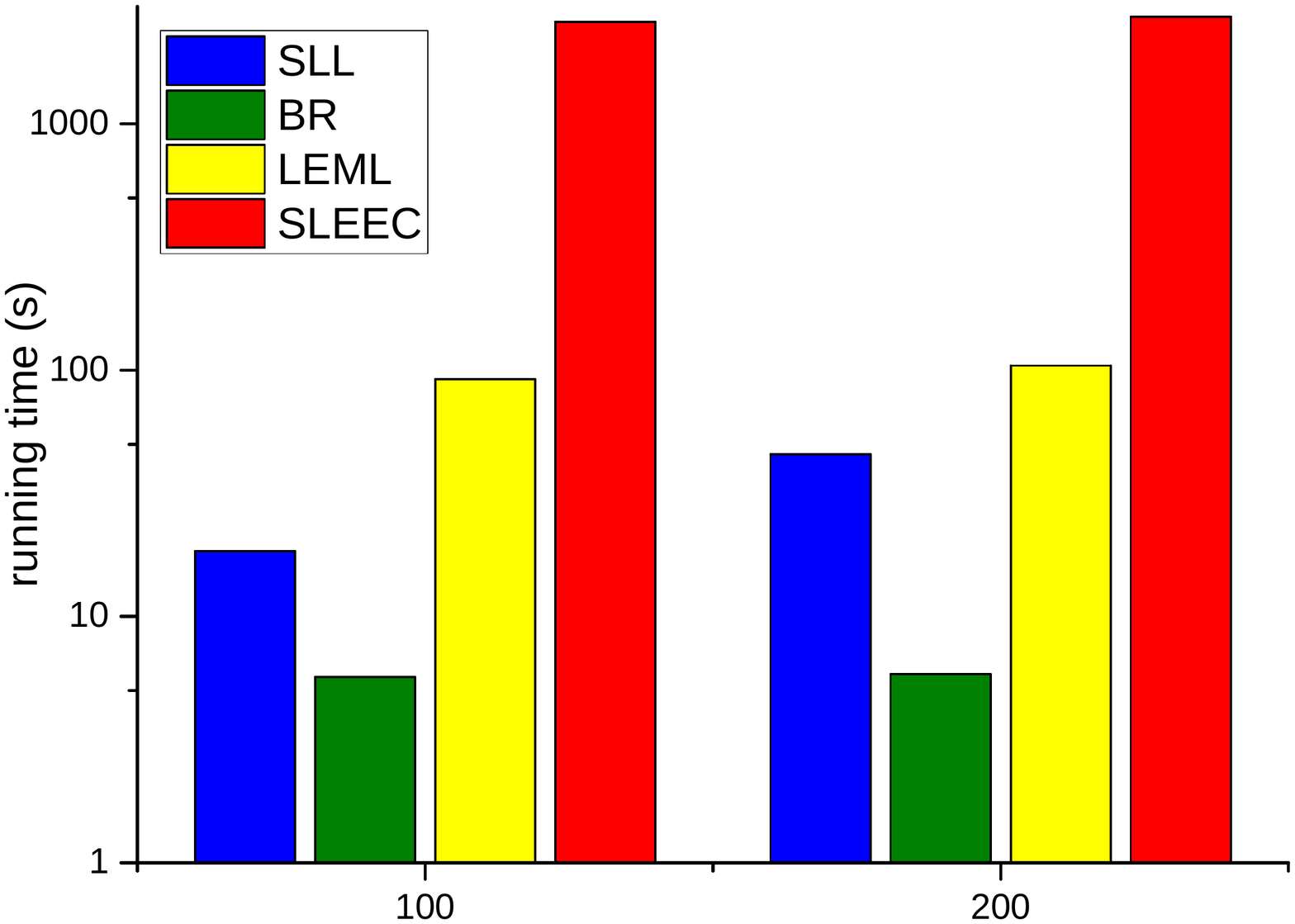}}
\hspace{3mm}
\subfloat[EURlex]
{\includegraphics[width=0.6\columnwidth]{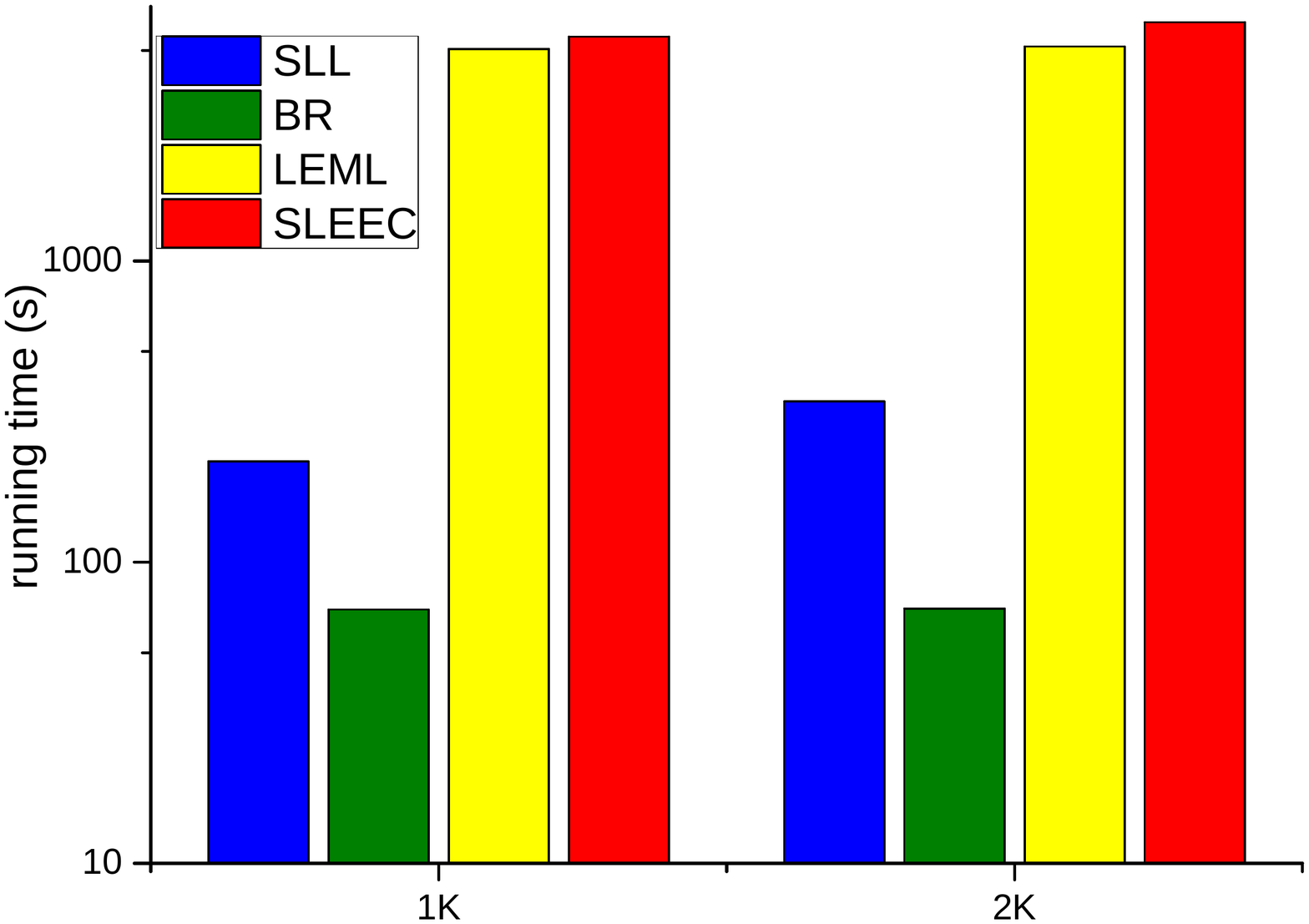}}
\hspace{3mm}
\subfloat[Wiki10]
{\includegraphics[width=0.6\columnwidth]{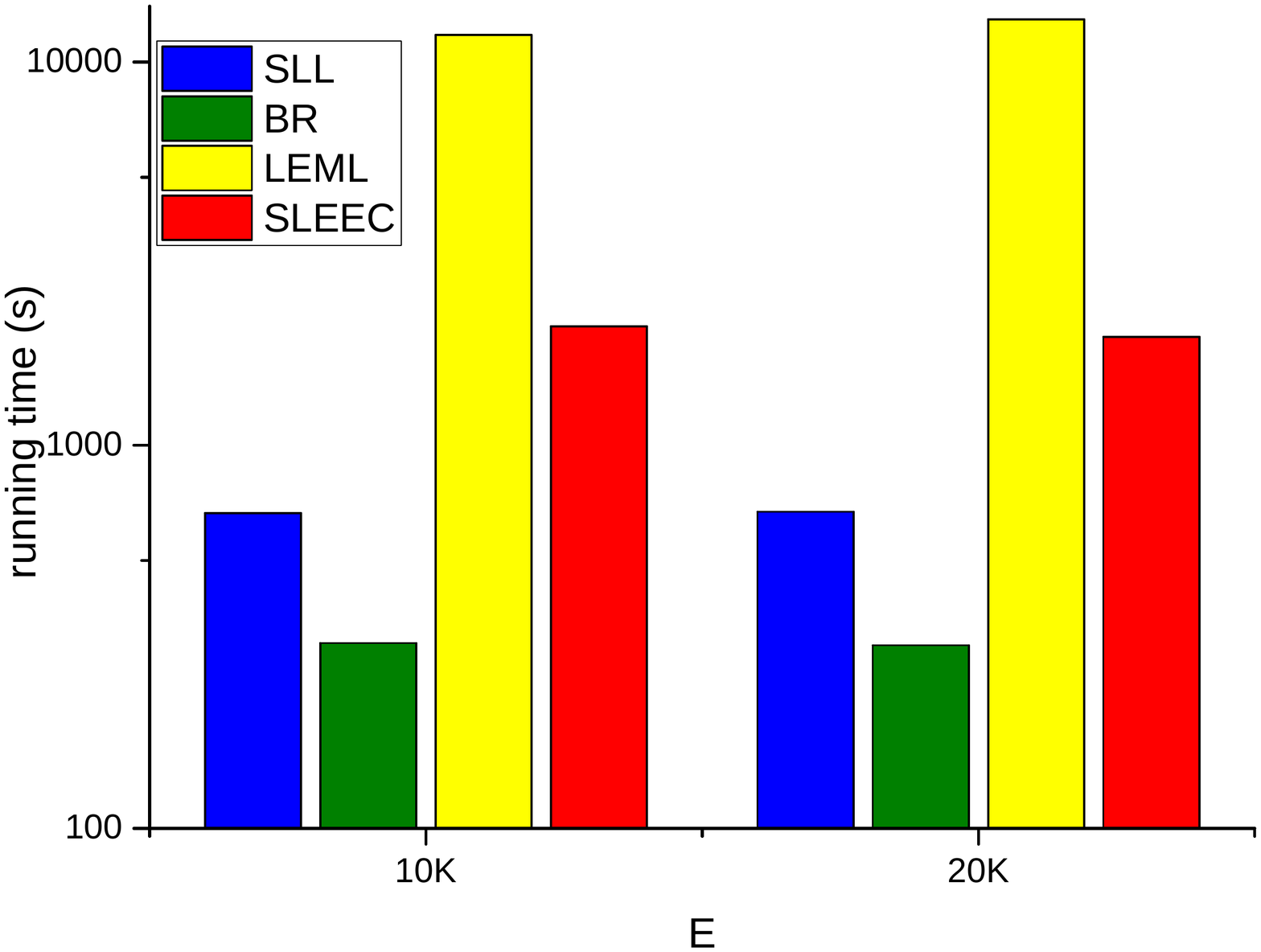}}
%\vskip -4mm
\caption{Running time of 100 new labels based on different number of past labels in large datasets.}
\label{time}
\end{figure*}

\begin{table*}% 加*表示跨栏，一般在双栏paper中使用
\caption{Comparison of learning new labels with different batch sizes by regarding 50\% labels as past labels.}
\label{biao2}
\centering
%\scriptsize
\footnotesize
%\tiny
%\small
\begin{tabular}{cr|rrrr||rrrr||rrrr}%
\multicolumn{2}{c|}{}& \multicolumn{4}{c||}{P@$1$} & \multicolumn{4}{c}{P@$3$} &\multicolumn{4}{c}{P@$5$}\\
\multicolumn{2}{c|}{}& SLL & BR & LEML & SLEEC        & SLL &BR& LEML & SLEEC  & SLL &BR& LEML & SLEEC     \\ \hline
\multirow{5}{*}{Bibtex} & 15 &\textbf{11.51} &10.85 &10.56 &10.89 &\textbf{5.09} &4.26 &4.12 & 4.45&\textbf{2.94} &2.68 &2.66 &2.70\\
                        & 30 &\textbf{18.20} &17.56 &17.48 &17.89 &\textbf{9.03} &8.42 &8.37 & 8.55&\textbf{5.52} &5.32 &5.27 &5.38\\
                        & 45 &\textbf{26.18} &25.25 &25.22 &25.84 &\textbf{12.98} &12.32 &12.29 & 12.55&\textbf{8.48} &8.03 &8.04 &8.19\\
                        & 60 &\textbf{41.11} &40.02 &40.16 &40.80 &\textbf{20.76} &19.78 &19.89 & 20.05&\textbf{13.62} &13.28 &13.32 &13.36\\
                        & 75 &\textbf{41.45} &40.91 & 40.97&41.31 &\textbf{22.85} &22.37 & 22.55& 22.61&\textbf{15.92} &15.29 &15.44 &15.53\\ \hline
\multirow{5}{*}{Mediamill} & 10 &\textbf{22.16} &21.36 &20.94 &21.75 &\textbf{13.62} &13.37 &13.32 & 13.50&\textbf{10.52} &10.39 &10.36 & 10.47\\
                        & 20 &\textbf{81.04} &80.27 &80.23 &80.31 &\textbf{53.41} &53.25 &53.21 & 53.36&\textbf{36.64}&36.51 &36.48 & 36.57\\
                        & 30 &\textbf{84.12} &54.16 &54.12 &83.99 &\textbf{54.26} &54.02 &54.01 & 54.20&\textbf{38.31} &38.06 &38.12 & 38.23\\
                        & 40 &\textbf{84.35} &84.12 &84.19 &84.27 &\textbf{55.35} &55.05 & 55.14& 55.28&\textbf{39.96} &39.68 &39.72 & 39.87\\
                        & 50 &\textbf{84.40} &84.22 &84.26 &84.34 &\textbf{55.88} &55.43 &55.66 & 55.80&\textbf{41.90} &41.45 & 41.78& 41.83\\ \hline
\multirow{5}{*}{Delicious} & 100 &\textbf{34.01} & 31.86&32.01 &32.12 &\textbf{23.04} &22.32&22.56& 22.74 &\textbf{18.16} &17.04 &17.22 & 17.32\\
                        & 200 &\textbf{38.97} & 38.03&38.18 &38.34 &\textbf{31.11} &30.01&30.17& 30.60 &\textbf{25.34} &24.42 &25.00 & 25.05\\
                        & 300 &\textbf{52.89} &52.13 &52.22 &52.78 &\textbf{43.47} &42.88&43.03& 43.30 &\textbf{36.09} &35.65 & 35.86& 35.97\\
                        & 400 &\textbf{58.52} &57.84 &58.00 &58.30 &\textbf{49.63} &48.78&49.22& 49.47 &\textbf{43.87} &43.08 &43.20 & 43.35\\
                        & 500 &\textbf{61.66} &61.23 &61.49 &61.57 &\textbf{54.52} &54.11&54.27& 54.41 &\textbf{48.50} &48.04 &48.12 & 48.33\\ \hline
\multirow{5}{*}{EURlex} & 200 &\textbf{5.12} &4.62 &4.69 & 4.81 & \textbf{2.09}   &1.86 &1.91 &  1.96   &\textbf{1.28}  & 1.09 &1.17 & 1.21  \\
                        & 400 &\textbf{11.39} &10.34 &10.55 & 10.70 &\textbf{4.95}    &4.22 &4.76 &  4.88   &\textbf{3.05}  &2.87  &2.96 & 3.01  \\
                        & 600 &\textbf{12.48} &11.81 &11.97 & 12.09 &\textbf{5.86}    &5.29 &5.54 &  5.77   &\textbf{3.68}  & 3.57 &3.62 & 3.66  \\
                        & 800 &\textbf{14.01} &13.23 &13.77 & 13.90 &\textbf{6.83}    &6.42 &6.66 &  6.72   &\textbf{4.26}  &4.02  &4.17 & 4.23  \\
                        &1000 &17.28&17.12 &17.26 & \textbf{17.31} &9.05&8.84 &9.02 &  \textbf{9.08}   &5.77  & 5.33 &5.65 & \textbf{5.79}  \\ \hline
\multirow{5}{*}{Wiki10} & 1k &\textbf{10.19} &9.23 &9.98 &10.02  &\textbf{4.82} &4.55&4.61&4.75  &\textbf{3.19} &3.11 &3.12 & 3.15\\
                        & 2k &\textbf{15.62} &15.28 &15.36 &15.54  &\textbf{7.95} &7.69&7.77&7.84  &\textbf{5.44} &5.28 &5.33 & 5.38\\
                        & 3k &\textbf{23.18}  &22.71 &22.85 &23.05  &\textbf{12.44} &12.22&12.30&12.38  &\textbf{8.63} &8.43 &8.51 & 8.56\\
                        & 4k &\textbf{31.11}  &30.22 &30.79 &31.00  &\textbf{18.07} &17.34&17.88&18.00  &\textbf{12.52} &12.33 &12.40 & 12.47\\
                        & 5k &38.69 &38.29 &38.55 &\textbf{38.71}  &\textbf{22.22} &21.78&22.10&22.21  &\textbf{15.40} &15.27 &15.32 & 15.37\\ \hline
\end{tabular}
\end{table*}

\noindent
\textbf{Mini-batch new labels.} For handling new labels in the mini-batch fashion, we investigate the results with different batch sizes. Specifically, for each dataset we randomly choose 50\% (for Delicious 483) labels as past labels since it tends to be a sensible option. Then we focus on the following new labels with different batch sizes. Instead of single new label senecio, in this case a batch of new labels can be independently processed within the traditional multi-label learning for LEML and SLEEC, including BR. The average results in various batches of labels are shown in Table \ref{biao2}. As indicated by the results, SLL largely outperforms other methods in dealing with mini-batch new labels. Under this setting, BR, LEML and SLEEC cannot employ the knowledge from past labels and regard the learning of new labels as an independent process. Nevertheless, SLL enables to learn new labels based on the obtained knowledge. Note that with the increase of batch size, the gap between SLL and other methods would shrink since the amount of labels is already large for well train a multi-label model; nevertheless, the training cost will increase accordingly.

%\begin{figure}[htbp]
%\centering
%\subfigure{\label{第1个子图标签名}}\addtocounter{subfigure}{-2}
%\subfigure[The 1st subfigure caption]{\subfigure[第1个子图标题]
%{\includegraphics[width=0.4\textwidth]{文件名}}}
%\subfigure{\label{第2个子图标签名}}\addtocounter{subfigure}{-2}
%\subfigure[The 2nd subfigure caption]{\subfigure[第2个子图标题]
%{\includegraphics[width=0.4\textwidth]{文件名}}}
%\bicaption[总标签名]{}{中文总标题}{Fig.$\!$}{The total caption}
%\vspace{-1em}
%\end{figure}

\subsection{Efficiency in dealing with new labels}
So far we have shown the superiority of SLL in classification performance, we now evaluate its efficiency in larger datasets. We select 100/200, 1k/2k and 10k/20k labels of Delicious, EURlex and Wiki10 respectively to serve initialization, then focus on the performance of 100 new labels. For SLL, we adopt the same initialization results with LEML. The average running time is showed in Figure \ref{time}, zoomed out using log10 scale. We can see that SLL and BR clearly surpass LEML and SLEEC in running time since they do not need the expensive retraining process. Note that the difference between SLL and BR lies in the label structure probe procedure, and for large datasets we can use clustering trick to reduce the scale of problem (\ref{new:weight}). For example, we select 3k labels for Wiki10 to form the fixed dictionary in Lasso, thus with the increase of label size, SLL can still be comparative with BR in efficiency.
%However, SLL has the competing classification performance with the state-of-the-art LEML and SLEEC, outperforming BR to a great extent since BR fails to capture the knowledge harvested from past labels.
%In addition, with the increasing size of past labels (100 to 200 in Delicious, 1k to 2k in EURlex, 10k to 20k in Wiki10), the time cost of SLL in learning new labels retains a slight increase while for LEME the cost is increasing dramatically. This indicates that the retraining process would be considerably time-consuming in terms of extremely large dataset, and our SLL can handle this case with efficient and accurate performance.

\subsection{Investigated label structure}
Since the adopted label structure plays an significant role in SLL because it helps to train classifiers with a special regularization, we intuitively show the investigated label relationships, \ie for a given label what labels are involved in its reconstruction. We select three labels and their five representation ``neighbors" of Bibtex dataset in Table \ref{structure}. As shown in Table \ref{structure}, we can easily see that some related labels in logic are exactly investigated by our SLL, which intuitively explains the reasons that SLL works. For example, to label ``epitope'', which is a terminology in immunology, some investigated labels are connected with its description, such as ``honogeneous" and ``sequence" while some are also in immunology, including ``lipsome" and ``immunosensor".

\begin{table}[ht]% 加*表示跨栏，一般在双栏paper中使用
\caption{Some related labels investigated by SLL in Bibtex.}
\label{structure}
\centering
\normalsize
\begin{tabular}{r|l}%
\hline
\multirow{2}{*}{epitope}& sequence; ldl ; homogeneous ; liposome ; \\
                        & immunosensor                                                   \\ \hline
\multirow{2}{*}{fornepomuk} & nepomuk ; langen ; knowledge ; semantics; \\
                             &  knowledgemanagement       \\ \hline
\multirow{2}{*}{concept}& formal; requirements; empirical; data; \\
                           & objectoriented  \\ \hline
\end{tabular}
\end{table}

\section{Conclusion}
In this paper we studied the streaming label learning (SLL) framework, which enables to model newly-arrived labels with the help of the knowledge learned from past labels. More precisely, we investigate the relationships among labels by examining label matrix from the perspective of label space and propose the label structure to embed it into the empirical risk minimization (ERM) framework, which regularizes the learning of the new classifier. We showed SLL can provide a tighter generalization bound of new labels and would not lose accuracy because SLL explores and exploits the label relationship. Thus SLL can be viewed as an efficient way to learn new classifiers under multi-label learning framework, but with no need of retraining the whole multi-label model. Experiments comprehensively demonstrated the superiority of SLL to existing multi-label learning methods in terms of handling new labels.

\appendices

\section{Proof of Theorem \ref{theorem:1}} \label{proof:1}
In this section, we detail the proof of Theorem \ref{theorem:1}, following the framework presented in \cite{yu2014large}. In the sequel, we will show that under streaming label learning, a tighter uniform convergence bound for the empirical losses will be obtained in terms of new classifiers. Denote the regularization set of SLL as $\mathcal{W}:= \{W\in\mathbb{R}^{d\times k},\norm{W-W_{old}S}_F \leq \varepsilon, \norm{S}_{1,1} \leq \lambda\}$, where $W_{old}$ is the previous multi-label classifier matrix for past labels and $S$ is the representation weight matrix of new labels with a sparsity inducing parameter $\lambda$.

The goal of the proof is to show with high probability the following inequality holds:
\begin{equation*}\label{proof:bound}
\mathcal{L}(\hat{W}) \leq \hat{\mathcal{L}}(\hat{W}) + \epsilon
\end{equation*}
where $\epsilon$ is a small quantity. $\mathcal{L}(W) = \Exp_{(\x,\y)} \qileft \ell(\y,f(\x;W))\qiright = \Exp_{(\x,\y)} \qileft \ell(\y,\x,W)\qiright = \Exp_{(\x,\y)} \qileft \sum_{l=1}^k\ell(\y^l,\x,\w_l) \qiright$ is the expectation of loss function $\ell$ or the \textit{real} loss. $\hat{\mathcal{L}}(W) = \frac{1}{n}\sum_{i=1}^n \ell(\y_i,\x_i,W) = \frac{1}{n}\sum_{i=1}^n \sum_{l=1}^k\ell(\y^l_i,\x_i,\w_l)$ is the empirical risk of the training data. Let $\displaystyle{W^* \in \arg\min_{W\in\mathcal{W}}\mathcal{L}(W)}, \displaystyle{\hat{W} \in \arg\min_{W\in\mathcal{W}}\hat{\mathcal{L}}(W)}$, then similar analysis can be implemented to obtain $\hat{\mathcal{L}}(W^*) \leq \mathcal{L}(W^*) + \epsilon$, inducing the ultimate inequality we claim, i.e. $\mathcal{L}(\hat{W})\leq\mathcal{L}(W^*) + 2\epsilon$. Thus we focus on the original uniform convergence bound. Typically, the whole proof of Eq.\ref{proof:bound} can be accomplished within three steps. We will elaborate them in the sequel.

\subsection{Bounding Excess Risk Using Its Supremum}
To probe an appropriate upper bound of the excess risk $\mathcal{L}(\hat{W})-\hat{\mathcal{L}}(\hat{W})$, it is natural to investigate its supremum,
\begin{equation*}
\begin{split}
&\mathcal{L}(\hat{W})-\hat{\mathcal{L}}(\hat{W}) \leq  \sup_{W\in\mathcal{W}} \{\mathcal{L}(W) - \hat{\mathcal{L}}(W) \} \\
= &\sup_{W\in\mathcal{W}}\left\{ \Exp_{(\widetilde{\x}_i,\widetilde{\y}_i)}\qiLeft\frac{1}{n}\sum_{i=1}^n \ell(\widetilde{\y}_i,\widetilde{\x}_i,W) \qiRight  - \frac{1}{n}\sum_{i=1}^n \ell(\y_i,\x_i,W)  \right\}\\
 \triangleq & g((\x_1,\y_1),...,(\x_n,\y_n))
\end{split}
\end{equation*}
For the decomposable loss function $\ell(\y,\x,W)= \sum_{l=1}^k\ell(\y^l,\x,\w_l) $, the change in any $(\x_i,\y_i)$ would induce a perturbation of $g((\x_1,\y_1),...,(\x_n,\y_n))$ at most $\mathcal{O}(\frac{k}{n})$. Then by using McDiarmid's inequality, the sum of squared perturbations will be bounded by $\frac{2k^2}{n}$, and thus the excess risk is bounded by a term related to the expectation of $g((\x_1,\y_1),...,(\x_n,\y_n))$, the expected supr\={e}mus deviation. Therefore, with probability at least $1-\delta$, it holds that
\begin{multline*}
\mathcal{L}(\hat{W})-\hat{\mathcal{L}}(\hat{W})\\ \leq \Exp_{(\x_i,\y_i)} \qileft g((\x_1,\y_1),...,(\x_n,\y_n)) \qiright + \mathcal{O}\left(k\sqrt{\frac{\log\frac{1}{\delta}}{n}}\right)
\end{multline*}
In the sequel, we will investigate the upper bound of the expected supr\={e}mus deviation.

\subsection{Bounding Expected Supr\={e}mus Deviation by a Rademacher Average}
Now we bound the expected supr\={e}mus deviation using its calculation and some tricks related to the  Rademacher complexity. Note that we adopt a Rademacher average introduced in \cite{yu2014large}. And we have
\begin{equation*}
\begin{split}
& \Exp_{(\x_i,\y_i)} \qileft g((\x_1,\y_1),...,(\x_n,\y_n)) \qiright\\
= &  {\normalsize \Exp\limits_{(\x_i,\y_i)}\qiLeft \sup_{W\in\mathcal{W}}\left\{ \Exp_{(\widetilde{\x}_i,\widetilde{\y}_i)}\qiLeft\frac{1}{n}\sum_{i=1}^n \ell(\widetilde{\y}_i,\widetilde{\x}_i,W)-\ell(\y_i,\x_i,W) \qiRight  \right\} \qiRight}  \\
\leq & \mathop{\Exp\limits_{(\x_i,\y_i)}}\limits_{(\widetilde{\x}_i,\widetilde{\y}_i)} \qiLeft \sup_{W\in\mathcal{W}}\left\{\frac{1}{n}\sum_{i=1}^n \ell(\widetilde{\y}_i,\widetilde{\x}_i,W)- \frac{1}{n}\sum_{i=1}^n \ell(\y_i,\x_i,W)  \right\}   \qiRight \\
 = & \mathop{\Exp_{(\x_i,\y_i)}}\limits_{(\widetilde{\x}_i,\widetilde{\y}_i),\epsilon_i} \qiLeft \sup_{W\in\mathcal{W}}\left\{\frac{1}{n}\sum_{i=1}^n \epsilon_i\left(\ell(\widetilde{\y}_i,\widetilde{\x}_i,W)- \ell(\y_i,\x_i,W)\right)  \right\}   \qiRight\\
\leq & \Exp_{(\x_i,\y_i),(\widetilde{\x}_i,\widetilde{\y}_i),\epsilon_i} \qiLeft \sup_{W\in\mathcal{W}}\left\{\frac{1}{n}\sum_{i=1}^n \epsilon_i\ell(\widetilde{\y}_i,\widetilde{\x}_i,W)\right\}\qiRight\\
&    +\Exp_{(\x_i,\y_i),(\widetilde{\x}_i,\widetilde{\y}_i),\epsilon_i} \qiLeft \sup_{W\in\mathcal{W}}\left\{\frac{1}{n}\sum_{i=1}^n \epsilon_i \ell(\y_i,\x_i,W)  \right\}   \qiRight\\
= &2 \Exp_{(\x_i,\y_i),\epsilon_i} \qiLeft \sup_{W\in\mathcal{W}}\left\{\frac{1}{n}\sum_{i=1}^n \epsilon_i \ell(\y_i,\x_i,W)  \right\}   \qiRight \\
= & 2\Exp_{(\x_i,\y_i),\epsilon_i} \qiLeft \sup_{W\in\mathcal{W}}\left\{\frac{1}{n}\sum_{i=1}^n \epsilon_i \sum_{j=1}^k \ell(\y_i^j,\x_i,\w_j)  \right\}   \qiRight\\
\leq & \frac{2C}{n} \Exp_{(\x_i,\y_i),\epsilon_i} \qiLeft \sup_{W\in\mathcal{W}}\left\{\sum_{i=1}^n \epsilon_i \sum_{j=1}^k \langle\x_i,\w_j\rangle  \right\}   \qiRight \\
= &\frac{2C}{n} \Exp_{\x_i,\epsilon_i} \qiLeft \sup_{W\in\mathcal{W}}\left\{\sum_{i=1}^n \sum_{j=1}^k \langle\epsilon_i\x_i,\w_j\rangle  \right\}   \qiRight = 2C\mathcal{R}_n(\mathcal{W})
\end{split}
\end{equation*}
where $\epsilon_i(i=1,...,n)$ are the Rademacher variables. The first inequality utilizes the Jensen inequality and the last inequality is based on the assumption that the loss function $\ell$ is bounded and $C$-Lipschitz. Here we adopt a Redemacher complexity defined as follows:
\begin{equation*}
\begin{split}
\mathcal{R}_n(\mathcal{W}) &\triangleq \frac{1}{n} \Exp_{\x_i,\epsilon_i} \qiLeft \sup_{W\in\mathcal{W}}\left\{\sum_{i=1}^n \sum_{j=1}^k \langle\epsilon_i\x_i,\w_j\rangle  \right\}   \qiRight \\
&= \frac{1}{n} \Exp_{\bx,\bm{\epsilon}} \qiLeft \sup_{W\in\mathcal{W}}\langle W,\bx_{\bm{\epsilon}}\rangle  \qiRight
\end{split}
\end{equation*}
where $\bx_{\bm{\epsilon}} \triangleq [\sum_{i=1}^n\epsilon_i\x_i,...,\sum_{i=1}^n\epsilon_i\x_i]$. Then in the last step, we directly calculate and estimate the Redemacher complexity.

\subsection{Estimating the Redemacher Complexity}
By applying the Cauchy-Schwarz inequality and the Triangle inequality of matrix norms, we obtain
\begin{equation*}
\begin{split}
&\frac{1}{n} \Exp_{\bx,\bm{\epsilon}} \qiLeft \sup_{W\in\mathcal{W}}\langle W,\bx_{\bm{\epsilon}}\rangle  \qiRight \leq \frac{1}{n} \Exp_{\bx,\bm{\epsilon}} \qiLeft \sup_{W\in\mathcal{W}}\norm{W}_F\norm{\bx_{\bm{\epsilon}}}_F  \qiRight\\
&= \frac{1}{n} \Exp_{\bx,\bm{\epsilon}} \qiLeft \sup_{W\in\mathcal{W}}\norm{W-W_{old}S+W_{old}S}_F\norm{\bx_{\bm{\epsilon}}}_F   \qiRight\\
  & \leq \frac{1}{n} \Exp_{\bx,\bm{\epsilon}} \qiLeft \sup_{W\in\mathcal{W}}(\norm{W-W_{old}S}_F+\norm{W_{old}S}_F)\norm{\bx_{\bm{\epsilon}}}_F   \qiRight\\
  &\leq \frac{\varepsilon+\norm{W_{old}S}_F}{n}\Exp_{\bx,\bm{\epsilon}} \qiLeft \norm{\bx_{\bm{\epsilon}}}_F   \qiRight
 \leq \frac{\varepsilon+\norm{W_{old}S}_F}{n}\sqrt{\Exp_{\bx,\bm{\epsilon}} \qiLeft \norm{\bx_{\bm{\epsilon}}}_F^2   \qiRight} \\
\end{split}
\end{equation*}
Then we calculate $\Exp_{\bx,\bm{\epsilon}} \qiLeft\norm{\bx_{\bm{\epsilon}}}_F^2\qiRight$ as,
\begin{equation*}
\begin{split}
\Exp_{\bx,\bm{\epsilon}} \qiLeft\norm{\bx_{\bm{\epsilon}}}_F^2\qiRight& = k\Exp_{\bx,\bm{\epsilon}} \qiLeft  \norm{\sum_{i=1}^n\epsilon_i\x_i}_2^2\qiRight \\
&= k \Exp_{\bx,\bm{\epsilon}} \qiLeft  \sum_{i=1}^n\norm{\x_i}_2^2 + \sum_{i\neq j}\epsilon_i\epsilon_j\langle\x_i,\x_j\rangle\qiRight \\
&= k\sum_{i=1}^n\Exp_{\x_i} \qileft\norm{\x_i}_2^2\qiright \leq kn
\end{split}
\end{equation*}
where the energy of feature $\Exp_{\x} \qileft\norm{\x}_2^2\qiright$ is assumed to be no more than 1 without loss of generality. Thus we can obtain an upper bound of the Redemacher complexity to establish Theorem \ref{theorem:1}:
\begin{equation*}
\begin{split}
\mathcal{R}_n(\mathcal{W}) &\leq \sqrt{\frac{k}{n}}(\varepsilon+\norm{W_{old}S}_F)\\
 &\leq \sqrt{\frac{k}{n}}(\varepsilon+\norm{W_{old}}_F\norm{S}_{1,1})\\
  &\leq \sqrt{\frac{k}{n}}(\varepsilon+\lambda\norm{W_{old}}_F).
\end{split}
\end{equation*}

\section{Proof of Theorem \ref{theorem:2}}\label{proof:2}
Now we detail the proof of Theorem \ref{theorem:2}, which aims to analyze the cost introduced by SLL, \ie, the perturbation of classifier parameter matrix (including both past labels and new labels) due to the SLL mechanism. Assuming the past label size is $m$ and a new label is given, the goal is to estimate the difference of the classifier matrix between the SLL and the original $m+1$ parameter matrix. Given the training data $\bx=[\x_1,...,\x_n]\in\mathbb{R}^{d\times n}$ and their label matrix $\by=[\y_1,...,\y_n]\in\{-1,1\}^{(m+1)\times n}$, the $m+1$-dimensional classifier parameter matrix is determined by the following optimization:
\begin{equation*}
\hat{Z} = \argmin_{Z\in\mathbb{R}^{d\times (m+1)}}J(Z) =  \sum_{i=1}^n \ell(\y_i,\x_i,Z) + \frac{\lambda}{2}\norm{Z-ZS}_F^2,
\end{equation*}
where $S$ is the label structure matrix of all $m+1$ labels.
%In the streaming label learning, the learning process is implemented in two consecutive steps:
%\begin{align*}
%&\hat{W} = \argmin_{W\in\mathbb{R}^{d\times m}} J_1(W) = \sum_{i=1}^n \ell(\y_i^{1:m},\x_i,W) + \frac{\lambda}{2}\norm{W-WS_m}_F^2,\\
%&\hat{\w} = \argmin_{\w\in\mathbb{R}^d} J_2(\w) = \sum_{i=1}^n \ell(y_i^{m+1},\x_i,\w) + \frac{\lambda}{2}\norm{\w-\hat{W}\s}_2^2,
%\end{align*}
Denoting $\tilde{Z} = [\hat{W},\hat{\w}]$, we have $J(\tilde{Z})\geq J(\hat{Z})$, and substitute it into the expression of $J(Z)$, and then we have
\begin{multline*}
\sum_{i=1}^n \ell(\y_i,\x_i,\tilde{Z}) - \sum_{i=1}^n \ell(\y_i,\x_i,\hat{Z})\\ \geq \frac{\lambda}{2}\norm{\hat{Z}-\hat{Z}S}_F^2 - \frac{\lambda}{2}\norm{\tilde{Z}-\tilde{Z}S}_F^2.
\end{multline*}
Denote the approximation error as $\Delta = \hat{Z} - \tilde{Z} = \hat{Z} - [\hat{W},\hat{\w}]$. Then the right hand side of the above inequality can be rewritten as
\begin{equation*}
\begin{split}
\mbox{right} =& \frac{\lambda}{2}\norm{\tilde{Z}+\Delta - (\tilde{Z}+\Delta )S}_F^2  - \frac{\lambda}{2}\norm{\tilde{Z}-\tilde{Z}S}_F^2\\
=&  \frac{\lambda}{2}\norm{\tilde{Z}-\tilde{Z}S}_F^2+\frac{\lambda}{2} \norm{\Delta-\Delta S}_F^2\\
 &+ \lambda Tr[(\tilde{Z}-\tilde{Z}S)^T(\Delta-\Delta S)]-\frac{\lambda}{2}\norm{\tilde{Z}-\tilde{Z}S}_F^2\\
=&  \frac{\lambda}{2} \norm{\Delta-\Delta S}_F^2 + \lambda \langle\Delta,(\tilde{Z}-\tilde{Z}S)(\bi-S)^T\rangle
\end{split}
\end{equation*}
where $\langle\cdot,\cdot\rangle$ is the inner product of two matrices. Similarly, the left hand side can also be rewritten using the approximation error $\Delta$ and for simplicity we consider the least squares loss function,
\begin{equation*}
\begin{split}
\mbox{left}= & \frac{1}{2}\norm{\by - \tilde{Z}^T\bx}_F^2 - \frac{1}{2}\norm{\by - \hat{Z}^T\bx}_F^2 \\
=& \frac{1}{2}\norm{\by - \tilde{Z}^T\bx}_F^2 - \frac{1}{2}\norm{\by - (\tilde{Z}+\Delta)^T\bx}_F^2\\
=&  \frac{1}{2}\norm{\by - \tilde{Z}^T\bx}_F^2 - \frac{1}{2}\norm{\by - \tilde{Z}^T\bx}_F^2\\
 &- \frac{1}{2}\norm{\Delta^T\bx}_F^2 + Tr[(\by - \tilde{Z}^T\bx)^T\Delta^T\bx]\\
=&  - \frac{1}{2}\norm{\Delta^T\bx}_F^2 + \langle\bx(\by - \tilde{Z}^T\bx)^T,\Delta\rangle
\end{split}
\end{equation*}
Thus we obtain
\begin{multline*}
\frac{\lambda}{2} \norm{\Delta-\Delta S}_F^2 + \frac{1}{2}\norm{\Delta^T\bx}_F^2 \\
\leq \langle\bx(\by - \tilde{Z}^T\bx)^T,\Delta\rangle - \lambda \langle\Delta,(\tilde{Z}-\tilde{Z}S)(\bi-S)^T\rangle
\end{multline*}
Suppose $d\ll n$ and $\bx$ is of full row rank, and denote its smallest singular value as $\sigma_1(\bx)$, then based on the singular value decomposition, we have
\begin{multline*}
\frac{\lambda}{2} \norm{\Delta-\Delta S}_F^2 + \frac{1}{2}\norm{\Delta^T\bx}_F^2\\ \geq \frac{\lambda}{2} \sigma_1^2(\bi-S)\norm{\Delta}_F^2 + \frac{1}{2}\sigma_1^2(\bx)\norm{\Delta}_F^2
\end{multline*}
Thus
\begin{equation*}
\begin{split}
& \frac{1}{2}[\lambda\sigma_1^2(\bi-S)+\sigma_1^2(\bx)]\norm{\Delta}_F^2\\
\leq& \langle\Delta,\bx(\by - \tilde{Z}^T\bx)^T-\lambda(\tilde{Z}-\tilde{Z}S)(\bi-S)^T\rangle\\
\leq& \norm{\Delta}_F \norm{\bx(\by - \tilde{Z}^T\bx)^T-\lambda(\tilde{Z}-\tilde{Z}S)(\bi-S)^T}_F\\
\leq& \norm{\Delta}_F \left(\norm{\bx}_F\norm{\by - \tilde{Z}^T\bx}_F+\lambda \norm{\tilde{Z}}_F\norm{\bi-S}_F^2\right)
\end{split}
\end{equation*}
Dividing both sides using $\norm{\Delta}_F $, we obtain
\begin{equation*}
\begin{split}
&\frac{1}{2}[\lambda\sigma_1^2(\bi-S)+\sigma_1^2(\bx)]\norm{\Delta}_F\\
\leq&  \norm{\bx}_F\norm{\by - \tilde{Z}^T\bx}_F+\lambda \norm{\tilde{Z}}_F\norm{\bi-S}_F^2\\
\leq& \sqrt{n\Omega}\norm{\by - \tilde{Z}^T\bx}_F+\lambda \norm{\bi-S}_F^2 \sqrt{\norm{\hat{W}}_F^2+\norm{\hat{\w}}_2^2}
\end{split}
\end{equation*}
and it is equivalent to
\begin{multline*}
\norm{\Delta}_F\leq \frac{2}{\lambda\sigma_1^2(\bi-S)+\sigma_1^2(\bx)}\cdot\\ \left(\sqrt{n\Omega}\norm{\by - \tilde{Z}^T\bx}_F+\lambda \norm{\bi-S}_F^2 \sqrt{\norm{\hat{W}}_F^2+\norm{\hat{\w}}_2^2}\right),
\end{multline*}
which completes the proof of Theorem \ref{theorem:2}.
\ifCLASSOPTIONcaptionsoff
  \newpage
\fi



%\bibliographystyle{IEEEtran}
%\newpage
\bibliographystyle{IEEEtran}
\bibliography{example_paper}
%\bibliography{icmlbib}
%\bibliographystyle{icml2016}

%\begin{thebibliography}{1}
%\end{thebibliography}


%\end{IEEEbiography}
%% insert where needed to balance the two columns on the last page with
%%\newpage


%\vfill


\end{document}